\documentclass[10pt,journal,compsoc]{IEEEtran}
\usepackage{generic}
\usepackage{cite}
\usepackage{amsmath,amssymb,amsfonts}
\usepackage{algorithmic}
\usepackage{graphicx}
\usepackage{bbm}
\usepackage{textcomp}
\usepackage{algorithmic}
\usepackage{algorithm}
\usepackage{booktabs}
\usepackage{multirow}
\usepackage{xspace}

\newcommand{\mname}{MetaVA\xspace}

 \makeatletter
 \let\NAT@parse\undefined
 \makeatother

\usepackage{hyperref}

\newcommand\FOREACH[1]{\FOR{\textbf{each} #1}}
\def\BibTeX{{\rm B\kern-.05em{\sc i\kern-.025em b}\kern-.08em
    T\kern-.1667em\lower.7ex\hbox{E}\kern-.125emX}}
\begin{document}
\title{ \mname: Curriculum Meta-learning and Pre-fine-tuning of Deep Neural Networks for Detecting Ventricular Arrhythmias based on ECGs }

\author{Wenrui Zhang, Shijia Geng, Zhaoji Fu, Linlin Zheng, Chenyang Jiang, and Shenda Hong$^{*}$

\thanks{Wenrui Zhang is with the Department of Mathematics, National University of Singapore, Singapore, 119077, Singapore (e-mail: zhangwenrui@u.nus.edu).}
\thanks{Shijia Geng is with the HeartVoice Medical Technology, Hefei, 230027, China (e-mail: gengshijia@heartvoice.com.cn).}
\thanks{Zhaoji Fu is with the School of Management, University of Science and Technology of China, Hefei, 230026, China, and HeartVoice Medical Technology, Hefei, 230027, China (e-mail: fuzj@mail.ustc.edu.cn).}
\thanks{Linlin Zheng is with the Department of Electrocardiography, The First Affliated Hospital of Anhui Medical University, Hefei, 230022, China (e-mail: zhenglinlin@ahmu.edu.cn).}
\thanks{Chenyang Jiang is with the Key Laboratory of Cardiovascular Intervention and Regenerative Medicine of Zhejiang Province, Department of Cardiology, Sir Run Run Shaw Hospital, Zhejiang University School of Medicine, Hangzhou, 310020, China (e-mail: cyjiang@zju.edu.cn).}
\thanks{Shenda Hong is with the National Institute of Health Data Science at Peking University, Peking University, and Institute of Medical Technology, Health Science Center of Peking University, Beijing, 100191, China (e-mail: hongshenda@pku.edu.cn). }

\thanks{$^{*}$Corresponding Author.}

}

\IEEEtitleabstractindextext{
\begin{abstract}
Ventricular arrhythmias (VA) are the main causes of sudden cardiac death. Developing machine learning methods for detecting VA based on electrocardiograms (ECGs) can help save people’s lives. However, developing such machine learning models for ECGs is challenging because of the following: 1) group-level diversity from different subjects and 2) individual-level diversity from different moments of a single subject. In this study, we aim to solve these problems in the pre-training and fine-tuning stages. For the pre-training stage, we propose a novel model agnostic meta-learning (MAML) with curriculum learning (CL) method to solve group-level diversity. MAML is expected to better transfer the knowledge from a large dataset and use only a few recordings to quickly adapt the model to a new person. CL is supposed to further improve MAML by meta-learning from easy to difficult tasks. For the fine-tuning stage, we propose improved pre-fine-tuning to solve individual-level diversity. We conduct experiments using a combination of three publicly available ECG datasets. The results show that our method outperforms the compared methods in terms of all evaluation metrics. Ablation studies show that MAML and CL could help perform more evenly, and pre-fine-tuning could better fit the model to training data.  
\end{abstract}

}

\maketitle
\section{Introduction}

Ventricular arrhythmias (VA), including ventricular tachycardia (VT) and ventricular fibrillation (VF), are the main causes of the sudden cardiac death worldwide \cite{HEDMAN200491}. These cause severe abnormal heartbeats in the ventricles, preventing the circulation of blood and exposing life to the dangers of sudden death. According to statistics from the American Heart Association, 377,763 people died due to sudden cardiac arrest in the USA in 2018; however, the rate of sudden cardiac death has decreased by 4.6\% (6.8\% versus 11.4\%) in the last 4 years \cite{AHAreport}. Accurate and timely detection of VA based on electrocardiogram (ECG) can help increase the rate of saving people's lives by 15.4\% - 19.2\% \cite{AHAreport}. 

ECG records the electric physiological activities of the cardiac muscle. It is one of the most commonly used non-invasive diagnostic tools for treating cardiac diseases. ECG signals can help detect VA by detecting abnormal ventricular electrical activities based on the morphology of the ECG waves. A manual analysis of ECG can be performed by certified cardiologists. However, this procedure is time-consuming, and the results interpreted by different cardiologists are likely to be inconsistent. In this situation, computer-aided VA detection algorithms, such as wavelet transformation \cite{2012A} and variation mode decomposition \cite{2016DetectionVMD}, has drawn increasing attention.

Traditionally, machine learning methods for VA detection involve a two stage pipeline of feature extraction and model building. Methods using only a single feature were proposed at first. Five previously existing VA detection algorithms using a feature were compared by Jekova et al.\cite{Jekova2000Comparison}. Amann et al. evaluated algorithms for VA detection by selecting data at intervals of one second without preselection \cite{2005Reliability}. Subsequently, some multi-feature methods were proposed to improve the detection performance. Li et al. used a genetic algorithm and support vector machines (SVMs) to select features \cite{li2014}. A novel VA detection algorithm that combined previously proposed parameters using SVM classifiers was proposed by Alonso-Atienza et al. \cite{AlonsoAtienza2015}. Irusta et al. proposed a high-temporal resolution algorithm to discriminate shockable from unshockable rhythms in adults and children and showed individual differences \cite{Irusta2012}. Mohanty et al. presented a method using variational mode decomposition-based features and a C4.5 classifier to detect VA \cite{2020Machine}. However, excessive manual intervention and the need for expert knowledge still render traditional methods as insufficient \cite{J2017Computer, 2007Errors, 2006Common}.

Recently, owing to the development of deep learning, deep neural networks, particularly convolutional neural network, have been widely used for ECG arrhythmia classification \cite{hong2020opportunities,2017Automated,jia2020personalized,HongXMLS19, FUJITA2019computer,MATHEWS2018anovel,dakun2020ecg,sabut2021detection}. It has been demonstrated experimentally that deep learning features are more informative and deep learning methods are superior to traditional methods for disease detection \cite{better,Hong_2019,HongWZWSLX17}. A neural network with weighted fuzzy membership functions was proposed to detect VF and VT \cite{2008Discrimination}. A fuzzy Kohonen neural network was proposed to discriminate different types of VA using multifractality \cite{FNN}. An abnormality framework combines an auxiliary classifier generative adversarial network and a residual network connected in parallel with a long short-term memory network to detect VA using an unbalanced dataset \cite{ganresnet}. Mahwish et al. proposed a method for transforming ECG signals into images and detecting VA using those images \cite{mahwish2021from}. Deep neural networks can help to automatically extract useful features and improve performance. Thus, manual intervention is not required for selecting the parameters when using the end-to-end model. 

Despite the success of the deep learning model, there are two challenges that existing deep learning methods cannot handle well.
\begin{itemize}
\item \textbf{Group-level diversity}. ECG signals differ between individuals \cite{2017Selection}. In other words, it is unclear whether what the model learned from one individual can be generalized to other individuals. This may be due to serveral reasons, such as the placement of channels, experimental protocols, and types of subjects, thus limiting their applicability in clinical environments \cite{MSSC, BOOSTANI201777}.
\item \textbf{Individual-level diversity}. In addition to the generalization ability for different individuals, the model may perform poorly in some stages while performing well in other stages. This is understandable because people perform different activities in different stages, which undoubtedly affect the intracardiac rhythm \cite{ecgsurvey}. Therefore, the model trained by certain time segments may perform poorly in other stages.
\end{itemize}

To address the aforementioned limitations of the existing deep learning approaches, we propose \mname, which is a meta-learning procedure that consists of two processes: pre-training and fine-tuning. The model is first pre-trained by using a large dataset, followed by fine-tuning over the new target dataset. First, we combine a meta-learning method, model agnostic meta-learning (MAML), and a curriculum learning (CL) strategy when pre-training the model. MAML is used to learn a model that can quickly adapt to new individuals and the CL strategy is introduced to afford an ``easy-to-hard'' learning order for MAML. This combined method can be regarded as an advanced transfer learning method. Furthermore, we adopt a modified fine-tuning method to improve the model performance in different stages. This modified fine-tuning method absorbs the thought of MAML, but aims at different samples rather than paying attention to various tasks. The results prove that our \mname model can achieve better performance than the other compared methods, yielding an overall improvement in terms of the evaluation metrics used. 

\begin{figure*}
    \includegraphics[scale = 0.57]{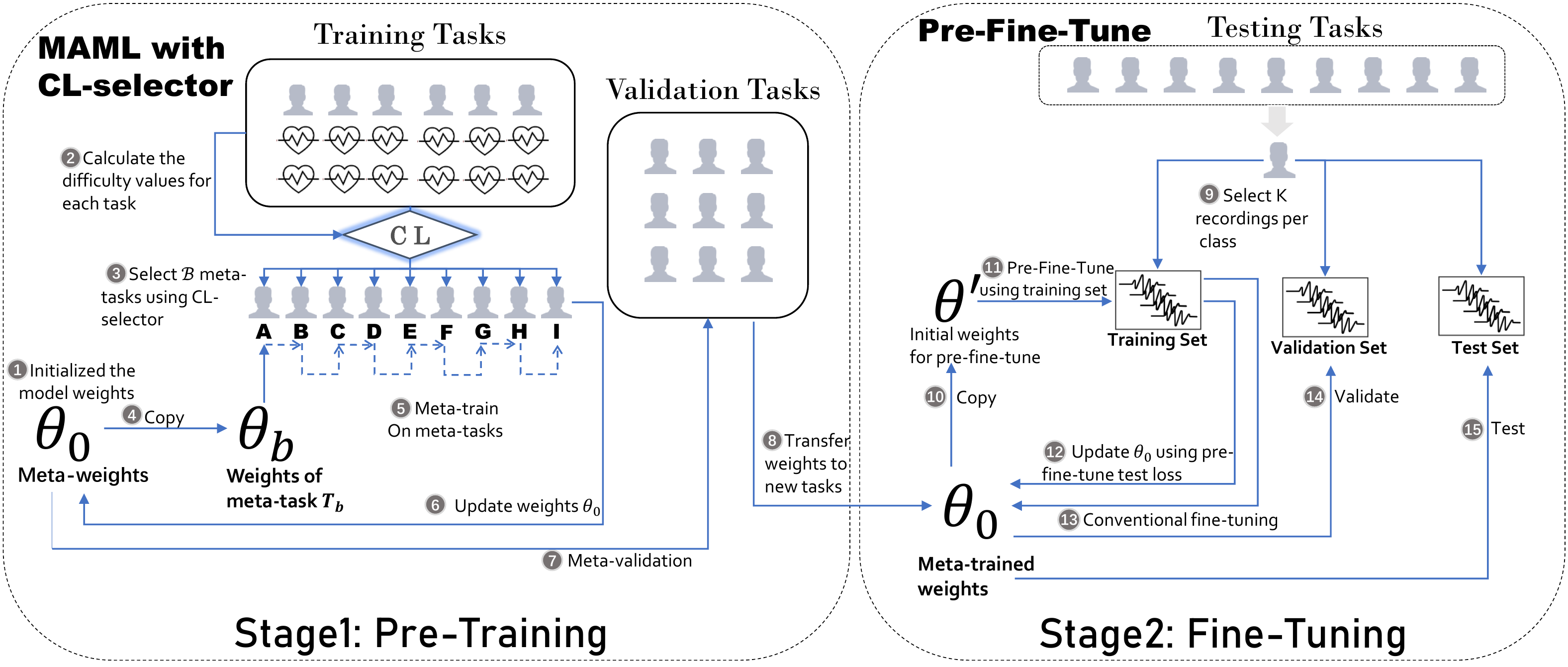}
    \caption{$\textit{MetaVADetector}$. Our proposed \mname model consists of two processes: pre-training and fine-tuning. In pre-training stage, we select different individuals as meta-tasks to meta-train the randomly initialized weights $\theta_0$. When the loss on validation tasks appears to increase, the pre-training stage ends and the trained $\theta_0$ is transferred to fine-tuning stage. In fine-tuning stage, we use recordings of each individual to train the $\theta_0$ to fit it to target person. We also use a validation set to choose hyperparameters and a test set to evaluate the performance.}
    \label{fig:process}
\end{figure*}

\section{Method}

In this section, we introduce our \mname method in detail.

As shown in \autoref{fig:process}, the entire \mname procedure consists of two stages: MAML with CL-selector (pre-training, Section \ref{wholeMD}) and pre-fine-tuning (fine-tuning, Section \ref{sec:finetune}). The pre-training stage aims to learn using large datasets and begins with randomly initialized weights $\theta_0$. The pre-training ends when the validation loss appears to increase and the weights $\theta_0$ are trained and transferred to the fine-tuning stage. The fine-tuning stage can be seen as an adaptation to new individuals and a test of the performance of $\theta_0$. A validation set is used to choose hyperparameters, and the performance is evaluated on a test set. The notations used in this study are listed in  \autoref{tab:notations}.

\begin{table}[h]
\caption{Notations}
    \centering
    \begin{tabular}{c|c}
    \toprule
    	Symbol & Description\\\midrule
        $\mathcal{T}$ &  Set of meta-training tasks\\
         $\mathcal{V}$&  Set of difficulty values \\
         $T_{train}$ & Set of meta-training tasks\\
         $V_{train}$ & Set of difficulty values of $T_{train}$\\
         $T_{val}$ & Set of meta-validation tasks\\
         $T_b$ & A meta-task\\
         $V_b$ & The difficulty value of $T_b$ \\
 	     $T_{new}$ & An unseen subject\\
         $\theta_0$ & Meta-weights\\
         $\theta_b$ & Copied parameters on meta-task $T_b$\\
         $f_\theta$ & Model parameterized by $\theta$\\
         $D_s$ & Support set of a meta-task\\
         $D_q$ & Query set of a meta-task\\
         $D_t$ & The training set of $T_{new}$\\
         $K$ & Number of recordings selected per class\\
         $B$ & Number of meta-tasks for meta-training\\
         $\mathcal{B}$ & Size of a mini-batch of tasks\\
     	\bottomrule
    \end{tabular}
    \label{tab:notations}
\end{table}

\subsection{Meta-detector with CL-selector}
\label{wholeMD}

\subsubsection{Meta Learning}
The field of meta-learning, also known as learn-to-learn, has gained increasing interest in recent years \cite{hospedales2020metalearning,vanschoren2018meta}. Meta-learning provides a method in which the trained model can gain experience over related tasks and improve its future learning performance using the experience \cite{hospedales2020metalearning}. MAML \cite{2017MAML} is a widely used meta-learning method that learns a set of pre-trained weights over multiple tasks. In contrast to traditional transfer learning which attempts to find the global optimum in all pre-trained tasks, MAML pays more attention to the potential of the parameters. In other words, MAML focuses on how to quickly adapt to the new target task and perform better on the new task. If we consider the disease detection for each person as a diverse task, MAML is an applicable method. In short, MAML can exhibit better generalization performance among tasks (persons) and fast adaptation, which serves our goal.

Formally, we denote that the neural network $f_\theta$ is parameterized by $\theta$. MAML attempts to find an initial set of parameters $\theta = \theta_0$ which could perform well on unseen tasks after only several stochastic gradient descent steps. $\theta_0$, also called meta-weights, is not globally optimum but has the greatest potential. Therefore, $\theta_0$ is supposed to be easily adaptable; that is, it can be quickly updated into a set of good $\theta_b$ on task $T_b$.

\subsubsection{Curriculum Learning}
CL \cite{09CurriculumLearning} represents the training strategies used to train the model from easy data to difficult data. This adopts the experience of human learning: learners are often offered the basic concepts first and gradually more advanced concepts later. We also want to organise the tasks in an ``easy-to-hard'' order. We aim to measure every meta-task's complexity or how difficult it is for the model to learn the tasks first.We consider the difficulty in learning each meta-task as the intrinsic trait of each task. In other words, if the model and model weights are determined, we can set a constant value for each task as its ``difficulty value''.  By determining the values in advance, we can obtain a more appropriate order to learn each meta-task without increasing much computational cost.

We can consider an easier CL strategy for a multitask problem. Assuming that the difficulty values of each task are $\mathcal{V} = \{V_0,V_1,\dots,V_B\}$ , we can set a selection function $sel(\mathcal{T},\mathcal{V}, \mathrm{i})$ to calculate the probabilities of choosing each task and perform a roulette wheel selection to choose the corresponding mini-batch of tasks at the i-th step. 
\begin{algorithm}
\caption{CL-selector initialization}
\label{alg:clselectorinit}
\begin{algorithmic}[1]
    \REQUIRE The set of training tasks $\mathcal{T}_{train}$
    \FOREACH{meta-task $T_b$ in $\mathcal{T}_{train}$}
        \STATE Trained using $K$ records of $T_b$
        \STATE Calculate the loss $loss_b$ on the other recordings of $T_b$
        \STATE Calculate the average loss : $loss_b \gets \frac{loss_b}{|T_b| - K}$
        \STATE Save $loss_b$ in $\mathbb{L}_{loss}$
    \ENDFOR

    \STATE Apply $Softmax'$ on $\mathbb{L}_{loss}$ using \autoref{equ:difficulty} and obtain $\mathcal{V}_{train}$
    \RETURN $\mathcal{V}_{train}$
    
\end{algorithmic}
\end{algorithm}
\subsubsection{Meta-detector with CL-selector}\label{sec:Pre-train}
We aim to employ MAML to achieve a fast adaptation of VA detection to new individuals. We regard each subject as a meta-task (denoted by $T_b$). After we pre-train (which is also called \textit{meta-train}) the model with a large VA dataset ($\mathcal{T} = \{T_1, T_2, \dots, T_B\}$) , we can obtain a potential set of parameters (meta-trained weights) $\theta_0$. During the pre-training stage, for every epoch, we employ a task-selector (referred to as CL-selector) to determine which meta-tasks to use in the current mini-batch. We consider that $\theta_0$ consolidates the knowledge of all the meta-tasks. When faced with a new unseen task $T_{new}$ , we can transfer the knowledge to $T_{new}$. The details of pre-training are described as follows.

We divide $\mathcal{T}$ into two sets: meta-training set ($T_{train}$) and meta-validation set ($T_{val}$).  Each subject in the dataset is treated as a single meta-task ($T_b$). The model weights are first initialized randomly using Xavier initialization \cite{2010Understanding}, which is a widely used randomized weight initialization method.\\

\begin{algorithm}
    \caption{CL-selector}
    \label{alg:clselector}
    \begin{algorithmic}[1]
    \REQUIRE Training tasks $T_{train}$, corresponding difficulty \\values set $V_{train}$, current meta-iteration \textit{iter}, batch-size of tasks $\mathcal{B}$, $MaxIter$ and $lowest$
    \STATE Calculate \textit{thres} : \\$thres = V_k < max(e^{iter-MaxIter}, lowest)$
    \FOR{b = 1 $\rightarrow$ $\mathcal{B}$}
    \FOREACH{meta-task $T_b$ in $T_{train}$}
        \STATE Calculate the probability $P_b$ of selecting $T_b$ using \autoref{equ:selectfunc}
        \STATE Retain $P_b + P_{b-1}$ ($P_b$ if b = 1) in $\mathbb{P}$
    \ENDFOR
    \STATE Generate a random number \textit{rand} in $[0,1]$
    \STATE Find index t where t first satisfies $\textit{rand} \leq \mathbb{P}[t]$
    \STATE Save t in $\mathbb{T}$
    \ENDFOR
    \RETURN $\mathbb{T}$
    \end{algorithmic}
\end{algorithm}

\begin{algorithm}
    \caption{MAML with CL-selector}\label{alg:MAML}
    \begin{algorithmic}[1] 
        \REQUIRE Meta-training tasks $T_{train}$
        \STATE Obtain $V_{train}$ using CL-selector initialization (\autoref{alg:clselectorinit})
        \FOREACH{meta-train iteration \textit{iter}}
        \STATE Select a mini-batch of training tasks: \\$T_{batch}$ = CL-selector($T_{train}$, $V_{train}$, \textit{iter}, $\mathcal{B}$) (\autoref{alg:clselector})
        	\FOREACH{meta-task $T_b$ in $T_{batch}$}
            	\STATE $\theta_b \gets \theta_0$
             	\STATE Randomly select $K$ epochs per class from $T_b$ for $D_s$ and $D_q$
                \FOR{\textit{step} = 1 $\rightarrow$ \textit{updates}}
                	\STATE Calculate loss on $D_s$ :$\mathcal{L}_{T_b}(f_{\theta_b})$
                    \STATE Update $\theta_b$ using \autoref{equ:inner_update}
                    \STATE Calculate loss on $D_q$ :$\mathcal{L}_{T_b}(f_{\theta_b})$
                    \STATE Retain $\mathcal{L}_{T_b}$ in  $\mathbb{L}_{T_b}$ : $\mathbb{L}_{T_b}.\textit{append}(\mathcal{L}_{T_b})$
                \ENDFOR
        	\ENDFOR
            \STATE Calculate sum loss of all meta-tasks using \autoref{equ:outer}
            \STATE Update weights $\theta_0$ using \autoref{equ:outer_update}
        \ENDFOR
    \end{algorithmic}
\end{algorithm}

\indent As described in \autoref{alg:MAML} and \autoref{fig:process}, we first calculate the difficulty values for each task, and can obtain $V_{train}$ . In detail, we train the model with weights $\theta_0$ using $K$ records in each meta-task $T_b$ in $T_{train}$ and test using the other records in $T_b$. Thus, for each $T_b$, we can obtain an average loss $loss_b$ of each recording. The difficulty value $V_b$ for each $T_b$ could be calculated after applying a function similar to the \textit{Softmax} function (referred to as $Softmax'$) on the losses:
\begin{equation}\label{equ:difficulty}
V_b = Softmax'(loss_b) =\frac{e^{loss_b}-1}{\sum_{i=1}^B(e^{loss_i}-1)}
\end{equation}
In each meta-training iteration, a mini-batch of $\mathcal{B}$ meta-tasks are selected by the CL-selector. The CL-selector can be regarded as a combination of function \textit{sel}() relative to the \textit{iter} (iteration), $t_k$ (times that $T_k$ has been selected) and $V_{k}$, and roulette wheel selection (selecting meta-tasks according to the probabilities). We obtain the probability that each meta-task should be selected via \textit{sel} and employ the roulette wheel selection. The selection function that we use is as follows:
\begin{equation}\label{equ:selectfunc}
sel(iter, t_k, V_k) = \frac{max(0,\mathbb{I}(V_k < \textit{thres}) -  \frac{t_k}{iter})}{\sum_{k=1}^B\mathbb{I}(V_k < \textit{thres})} 
\end{equation}\begin{equation*}
thres = max(V_k < e^{iter - MaxIter},\textit{lowest})
\end{equation*}
where $\mathbb{I}$ is the indicator function:
\begin{equation*}
    \mathbb{I}(x<a) = \left\{\begin{matrix}1, &x<a
\\ 0, &x\ge a

\end{matrix}\right.
\end{equation*},\textit{lowest} is the $\mathcal{B}$-th smallest  difficulty value, and \textit{MaxIter}, representing when all tasks can be selected, is set to 50. When calculating $V_b$, we use $Softmax'$ to amplify the relative values among the losses. When we select the meta-task, we simply set a threshold relative to \textit{iter} and select the tasks whose difficulty values are less than the threshold. Moreover, we reduce the possibility of selecting the tasks used before. The entire CL-selector is summarized in \autoref{alg:clselectorinit} and \autoref{alg:clselector}.

After selecting the mini-batch of tasks, the model is meta-trained using all the tasks in this mini-batch. To ensure the same start weights, each $T_b$ should copy the weights from $\theta_0$ as its own initial weights ($\theta_b$).

Subsequently, the support set $D_s$ (for training), and query set $D_q$ (for test), are randomly selected from $T_b$, and each of these consists of $K$ records for VA and non-VA. Thus, a total of  $K \times 2$ recordings are selected per set. Then, we perform gradient descent separately for \textit{update} steps. In each step, we first use $D_s$ to update $\theta_b$ as follows:
\begin{equation}\label{equ:inner_update}
\theta_b \gets \theta_b - \alpha\nabla_{\theta_b}\mathcal{L}_{T_b}(f_{\theta_b})
\end{equation}where $\alpha$ is the learning rate (we call it \textit{update\_lr} to distinguish the following \textit{meta\_lr}). Next, we evaluate the updated weights $\theta_b$ using the query set $D_q$, and obtain the loss $\mathcal{L}_{T_b}(D_q)$ which is retained in a list $\mathbb{L}_{T_b}$. After all $\mathcal{B}$ meta-tasks selected are completed, we can obtain the summation of the losses on $\mathcal{B}$ query sets:
\begin{equation}\label{equ:outer}
\mathcal{L}_{meta}(\theta_0) = \sum_{b=1}^{\mathcal{B}}\mathbb{L}_{T_b}(f_{\theta_b})
\end{equation}This function can be seen as the objective function for meta-training; that is, our goal is to minimize this equation. To update $\theta_0$ , we perform gradient descent of $\mathcal{L}_{meta}$  using
\begin{equation}\label{equ:outer_update}
\theta_0 \gets \theta_0 - \gamma\nabla_{\theta_0}\mathcal{L}_{meta}(\theta_0)
\end{equation}where $\gamma$ is the learning rate of meta-training (i.e., \textit{meta\_lr} as mentioned earlier).

In the meta-validation stage, we adapt the model to all meta-tasks in $T_{val}$ and test the performance. This procedure is applied to tune the hyperparameters, such as learning rates and \textit{updates}. We select $\theta_0$ when it achieves the best validation loss after adapting to the new validation meta-tasks.

\subsection{Pre-fine-tuning}\label{sec:finetune}

This stage is used to personalize the model to the new target subject. Instead of directly employing the traditional adaptation method in transfer learning, we add a method similar to MAML called pre-fine-tune before applying the traditional adaptation method, to overcome the problem that recordings in different stages could be significantly different. We know that MAML trains the model using the loss on the query set of the meta-weights (which have been updated) to adapt the model to different tasks. Similarly, we expect to train the model using the loss on the training set of copied and trained weights as well, to adapt the model to various recordings. We initialize the model weights with the pre-trained weights $\theta_0$, to transfer the knowledge of abundant data to this stage. Records from the unseen subject (denoted as $T_{new}$) are divided into three categories: $K$ records per class as a training set $D_t$, $K$ records per class as a validation set, and the remaining records as a test set.

With regard to the training set, we first use the proposed pre-fine-tune method to search a more appropriate neighborhood for the distinct records, as described in \autoref{alg:finetune}. Similar to MAML, we first copy weights $\theta'$ from the pre-trained $\theta_0$ and train $\theta'$ on the training set $D_t$, and test on the same set to obtain the loss $\mathcal{L}_{D_t}(f_{\theta''})$.
\begin{equation}\label{equ:gettheta''}
\theta'' \gets \theta' - \beta_1\nabla_{\theta'}\mathcal{L}_{D_t}(f_{\theta'})
\end{equation} where $\beta_1$ is the learning rate and $\mathcal{L}_{D_t}(f_{\theta'})$ is the loss on $D_t$ of weights $\theta'$. Then, we perform the gradient descent of $\mathcal{L}_{D_t}(f_{\theta''})$ on $\theta_0$:
\begin{equation}\label{equ:updatetheta0}
\theta_0 \gets \theta'' - \beta_2\nabla_{\theta''}\mathcal{L}_{D_t}(f_{\theta''})
\end{equation}where $\beta_2$ is the learning rate.\\
\indent After pre-fine-tuning, we continue with traditional fine-tuning and perform stochastic gradient descent on the individual subject. Traditional fine-tuning also trains the model on the same training set. Corresponding to finding a neighborhood by pre-fine-tuning, fine-tuning attempts to locally search a set of more accurate weights. In this stage, we expect to adapt the model to the new target using only a small amount of data; thus, $K$ is not a large number. $\theta_0$ is then validated on the validation tasks to find suitable hyperparameters. Finally, we test the model on the test tasks to evaluate the performance of weights $\theta_0$.

\begin{algorithm}
    \caption{Pre-Fine-Tune}\label{alg:finetune}
    \begin{algorithmic}[1] 
        \REQUIRE Training set $D_t$
        \FOREACH{pre-fine-tune iteration}
        	\STATE Copy the pre-trained weights : $\theta' \gets \theta_0$
        	\STATE Calculate loss of $\theta'$ : $\mathcal{L}_{D_t}(f_{\theta'})$
        	\STATE Get $\theta''$ using \autoref{equ:gettheta''}
        	\STATE Calculate loss of $\theta''$ : $\mathcal{L}_{D_t}(f_{\theta''})$
            \STATE Update $\theta_0$ using \autoref{equ:updatetheta0}
        \ENDFOR
    \end{algorithmic}
\end{algorithm}

\begin{table*}[ht]
\caption{Summary of Databases}\label{table:databases}
\centering
\begin{tabular}{l|cccc}
\toprule
Database & \# of Recordings & \# of Leads & Sampling Rate (Hz) & Recording Duration (minutes) \\ \midrule
MITDB    & 48                   & 2               & 360               & 30                      \\
VFDB     & 22                   & 2               & 250               & 30                      \\
CUDB     & 35                   & 1               & 250               & 8                      \\
\bottomrule
\end{tabular}
\end{table*}

\section{Experiments}
To verify the feasibility of employing MAML, the CL-selector, and pre-fine-tune in this task, we conduct experiments to support our methods as follows. First, we believe that using MAML can result in better initial weights than traditional pre-training methods. Second, the CL-selector could help MAML find more appropriate parameters. Finally, our pre-fine-tune could improve the performance of traditional fine-tuning when we made the model adapt to unseen persons. In all procedures, we conduct the two-part experiments: pre-training and fine-tuning. The experimental setup is described in this section.

\subsection{Datasets}
In the experiments, we combine the following three publicly available and widely used ECG datasets:
\begin{itemize}
\item MIT-BIH arrhythmia database (MITDB) \cite{mitdb1,mitdb2}. MITDB was set-up by the Beth Israel Hospital Arrhythmia Laboratory between 1975 and 1979, containing 48 30-minute recordings obtained from 47 subjects. Each ECG data point has two leads and the sampling rate is 360 Hz. 23 recordings were randomly selected and other 25 recordings represent less common but clinically important phenomenons. This database is the first publicly available set to evaluate arrhythmia detectors. 

\item MIT-BIH malignant ventricular arrhythmia database (VFDB) \cite{vfdb1,vfdb2}. VFDB was collected from the ECG tape libraries of the Brigham and Woman's Hospital and Beth Israel Hospital in Boston in 1986, containing 22 half-hour recordings obtained from 16 patients. Each data point has two leads as well, and the sampling rate is 250 Hz. This database is also widely used in VA detection tasks.

\item Creighton University ventricular tachyarrhythmia database (CUDB) \cite{cudb1,cudb2}. CUDB was originally collected by Floyd M. Nolle at the Creighton University Cardiac Center. It contains 35 eight-minute ECG recordings with sustained VT, VF, and ventricular flutter. All signals are digitized at 250 Hz with 12-bit resolution over a 10 V range, and each data point has only one lead. Such high-quality recordings of these rhythms are of great importance in VA detection tasks.

\end{itemize}

A summary of these three databases is presented in \autoref{table:databases}. 

The following pre-processing steps are performed: (1) lead \textit{MLII} is extracted from the MITDB and lead \textit{ECG} from the CUDB and VFDB. Other leads are ignored because of the small amount (such as lead \textit{V2}). (2) All recordings are re-sampled to 200 Hz using linear interpolation. (3) Long recordings are split into 2-second segments using a sliding window with dynamic strides. (4) Every segment is labeled as 1 (if VA exists) or 0 (if not VA). 
In addition, to deal with the imbalance problem, the stride is set to 20 (0.1 s $\times$ 200 Hz) if VA exists and 400 (2 s $\times$ 200 Hz) if not. The details of dividing the training, validation and test sets are provided in Section \ref{sec:experiments}, along with the method details.

\subsection{Implementation Details of Deep Neural Network}
Our method has good generalizability. It does not require any specific form of deep neural networks. Hence, the purpose of the experiments is not to compare different network architectures. Therefore, one neural network architecture is used for all the compared methods.

The model is modified based on \cite{hong2020holmes}, which presents a deep neural network backbone that achieves state-of-the-art performance in ECG modeling \cite{HongWZWSLX17,Hong_2019,hong2020holmes}. This can be seen as a structurally designed \cite{radosavovic2020designing} one-dimensional ResNeXt \cite{ResNext} model. A detailed model architecture is shown in \autoref{table:network}. Our employed network contains 44 layers in total and seven stages, each of which contains two blocks. The basic block is a bottleneck architecture, consisting of a cascade of three convolutional layers, and blocks are residual-connected with shortcut connections \cite{he2016Identity,he2016Deep}. The first and third convolutional layers are convolutional layers with the kernel size set to 1, and the second is an aggregated convolutional layer \cite{ResNext} with the kernel size set to 16 and groups set to 16. The output sizes of each stage are [128,64,64,32,32,16,16], and the input length is downsampled to half in the second layer of the first block in the 2-nd, 4-th and 6-th stages. The corresponding shortcut connections also use Max Pooling to downsample the identity. Before each convolutional layer, Swish activation \cite{Ramachandran2017Searching} and dropout \cite{2014Dropout} are employed to achieve a nonlinear transformation. After seven stages, the last dimension is pooled by an average layer, and the prediction layer is a fully connected dense layer.

\begin{table}[h]
\centering
\caption{Our network architecture}
\label{table:network}
\begin{tabular}{|c|c|c|}
\hline
Layer Name                       & Output Size          & Composition           \\ \hline
\multirow{2}{*}{Conv1}           & \multirow{2}{*}{16}& Conv1d                \\ \cline{3-3} 
                                 &                      & BN, Swish             \\ \hline
\multirow{2}{*}{Stage1}          & 16                  & Block1                \\ \cline{2-3} 
                                 & 16                  & Block2                \\ \hline
\multirow{2}{*}{Stage2}          & 32                   & Block1                \\ \cline{2-3} 
                                 & 32                   & Block2                \\ \hline
\multirow{2}{*}{Stage3}          & 32                 & Block1                \\ \cline{2-3} 
                                 & 32                   & Block2                \\ \hline
\multirow{2}{*}{Stage4}          & 64                   & Block1                \\ \cline{2-3} 
                                 & 64                   & Block2                \\ \hline
\multirow{2}{*}{Stage5}          & 64                   & Block1                \\ \cline{2-3} 
                                 & 64                   & Block2                \\ \hline
\multirow{2}{*}{Stage6}          & 128                   & Block1                \\ \cline{2-3} 
                                 & 128                  & Block2                \\ \hline
\multirow{2}{*}{Stage7}          & 128                  & Block1                \\ \cline{2-3} 
                                 & 128                  & Block2                \\ \hline
\multicolumn{1}{|l|}{Prediction} & 2                    & Fully Connected Layer \\ \hline
\end{tabular}
\end{table}

\subsection{Implementation Details of \mname and Comparisons}
\label{sec:experiments}

\begin{table*}
    \centering
    
    \caption{The combinations represented by compared methods and ours}
    \begin{tabular}{l|cc}
    \toprule
                   & Pre-Training Stage & Fine-Tuning Stage \\\midrule
        \mname & MAML with CL-selector & Pre-Fine-Tune \\
        MC+F & MAML with CL-selector & Fine-Tune\\
        M+PF & MAML  & Pre-Fine-Tune\\
        M+F & MAML & Fine-Tune\\
        Vanilla & Pre-Training & Fine-Tune\\
    \bottomrule
    \end{tabular}
    \label{table:combinations}
\end{table*}

\begin{table*}[ht]
\caption{Results of Combinations}\label{table:results1}
\centering
\setlength{\tabcolsep}{3.5mm}{
\begin{tabular}{l|ccccccc}
\toprule
Method  & ROC-AUC  & Accuracy & PR-AUC & F1-score \\ \midrule
\mname   & \textbf{0.9843$\pm$0.0041}&\textbf{0.9602$\pm$0.0060}&\textbf{0.9681$\pm$0.0126}&\textbf{0.9173$\pm$0.0057}   \\
MC+F& 0.9742$\pm$0.0061&0.9549$\pm$0.0089&0.9512$\pm$0.0218&0.9064$\pm$0.0150\\
M+PF     &0.9823$\pm$0.0103&0.9572$\pm$0.0076&0.9588$\pm$0.0248&0.9156$\pm$0.0076\\
M+F &0.9703$\pm$0.0093&0.9497$\pm$0.0132&0.9423$\pm$0.0185&0.8902$\pm$0.0171\\
Vanilla &0.9566$\pm$0.0056&0.9392$\pm$0.0141&0.9025$\pm$0.0269&0.8833$\pm$0.0181\\

\bottomrule
\end{tabular}
}
\end{table*}
In this section, we describe the implementation details and comparisons of the \mname model. For a better understanding, we can follow the basic pipeline in Figure \ref{fig:process} and then describe pre-training and fine-tuning separately. Then, the combinations of methods in the two parts are summarized in \autoref{table:combinations}.

\subsubsection{Pre-training}\label{sec:exppre}
As mentioned in Section \ref{sec:Pre-train}, $\mathcal{B}$ meta-tasks are selected by the CL-selector in every meta-training iteration, and $K$ recordings per class are randomly selected for training in each meta-task $T_b$, i.e., each subject. To achieve fast adaptation, the value of $K$ should not be too large and is set as 10. The individuals in the MITDB and CUDB are selected for meta-training and meta-validation, and the individuals in VFDB are used for fine-tuning and testing. Two methods of pre-training are performed to compare the performance of our approach. The details of each method are as follows:
\begin{itemize}
\item \textbf{MAML with CL-selector}. The meta-training procedures are explained in detail in \autoref{sec:Pre-train}. The number of meta-tasks selected by the CL-selector in each iteration ($\mathcal{B}$) is set to 9. For meta-validation, nine subjects are randomly selected from the MITDB (referred to as MITDB\uppercase\expandafter{\romannumeral2} and the remaining MITDB is referred to as MITDB\uppercase\expandafter{\romannumeral1}) and fixed in every model run. The model does not stop meta-training until an increasing trend of the loss of meta-validation appears. The model's weights $\theta_0$ are maintained at the best iteration. The hyperparameters include the learning rate \textit{update\_lr} ($\alpha$) $\in$ $\{10^{-2}, 5*10^{-3}, 10^{-3}\}$, \textit{meta\_lr} ($\gamma$) $\in \{10^{-2}, 5*10^{-3}, 10^{-3}\}$, and the number of updating steps (\textit{updates} $\in\{1, 3,5\}$).
\item \textbf{MAML}. MAML with randomly selected meta-tasks is chosen as a comparison method to test the hypothesis that the CL-selector could improve the performance of MAML. In each meta-iteration, $\mathcal{B}$ meta-tasks are selected randomly rather than by the CL-selector. This is the only difference between our method and this method. The set of hyperparameters also includes \textit{update\_lr} $\in \{10^{-2}, 5*10^{-3}, 10^{-3}\}$, \textit{meta\_lr} $\in \{10^{-2}, 5*10^{-3}, 10^{-3}\}$, and the updating steps \textit{updates} $\in\{1,3,5\}$. 
\item \textbf{Direct Pre-training}. The traditional pre-training approach is used in comparison with MAML. To directly compare this method with our method, the same number of training recordings as in our method in each iteration is selected. Because our method requires two sets ($D_q$ and $D_s$) when meta-training, $\mathcal{B} \times K \times 2$ recordings per class from each subject are selected to train the model to maintain the same number of training samples. Moreover, the training, validating, and testing tasks are the same as in our method. In every training iteration, the model performs gradient descent using mini-batches. The model is also trained until the validation loss does not descend and is maintained at its best iteration. The set of hyperparameters includes the learning rate $\in \{10^{-2}, 5*10^{-3}, 10^{-3}\}$, and batch size $\in\{ 32, 64, 128\}$.

\end{itemize}

All combinations of hyperparameters are tried, and the set that achieves the lowest validation loss is retained. The model with the best set of hyperparameters is trained five times, and five sets of weights $\theta_0$ of each method are saved to adapt to unseen subjects.
\subsubsection{Fine-tuning}\label{sec:expfinetune}
The knowledge from pre-training ($\theta_0$) is transferred to this stage to quickly adapt to the new subject. In this stage, the VFDB is used, and the only lead, called \textit{ECG}, is selected. For each subject in the VFDB, only $K=10$ recordings per class are selected to verify our hypothesis that MAML could help adapt to new individuals quickly. Moreover, other $K=10$ recordings per class are selected as validation sets to choose the appropriate set of hyperparameters. When initializing the model weights, three types of weights are used to compare our pre-training method (MAML with the CL-selector) with other methods: 1) $\theta_0$ from MAML with the CL-selector, 2) $\theta_0$ from MAML, and 3) $\theta_0$ from the traditional pre-training method. After initializing the model weights, we adopt traditional fine-tuning with pre-fine-tuning (see Section \ref{sec:finetune}) as our method and select traditional fine-tuning as the compared method. To make the results more convincing, the model runs 10 times per subject per weight initialization, and the recordings used for training and validation in the two methods are found to be all identical. The details are as follows.
\begin{itemize}
\item \textbf{Pre-fine-tuning} is described in Section \ref{sec:finetune}. After pre-fine-tuning, the gradient descent method is applied to the model until the training loss tends to be stable. The fine-tuned model is then tested on the remaining recordings. The set of hyperparameters includes the number of pre-fine-tuning iterations (\textit{iter} $\in \{10,30,50\}$), two learning rates in pre-fine-tuning ($\beta_1,\  \beta_2 \in \{10^{-2}, 5\times10^{-3},10^{-3}, 5\times10^{-4}\}$), and the learning rate in fine-tuning (\textit{learning\_rate}$\in\{10^{-2}, 10^{-3},10^{-4}\}$). The maximum number of training iterations in fine-tuning is set to 200.
\item \textbf{Direct fine-tuning} is employed to prove that pre-fine-tuning can improve the fine-tuning performance. As a result, there is no difference between fine-tuning in the aforementioned method and this method. The set of hyperparameters includes only \textit{learning\_rate}$\in \{10^{-2}, 10^{-3},10^{-4}\}$, and the maximum number of training iterations is still set to 200.
\end{itemize}

Finally, the compared methods are consisted by combining two parts together. The combinations are listed in \autoref{table:combinations}. Notice that MC+F is an advanced edition (added with a CL-selector) of MetaSleepLearner\cite{metasleeplearner}, which our MAML method is inspired by.

\subsection{Evaluations}
The performance metrics we selected to evaluate the performance of different combinations are the receiver operating characteristics area under the curve (ROC-AUC), precision-recall area under the curve (PR-AUC), accuracy, and F1-score. The x-axis of the ROC curve is the false positive rate, and the y-axis is the true positive rate. The AUC can measure a binary classifier system as the discrimination threshold varies. It can also be regarded as the probability that a positive example ranks higher than a negative example when randomly chosen\cite{AUC}. PR-AUC represents the mean precision under different discrimination thresholds. The F1-score and accuracy are relative to the discrimination threshold. The F1-score is the harmonic mean of the precision and recall. $$\mathrm{F_1} = 2 \times \frac{precision \times recall}{precision + recall}$$. Accuracy is the ratio of the number of truly classified samples to the number of samples. For the accuracy and F1-score, we obtain the threshold for classifying labels by maximizing the geometric mean value\cite{gmean}. 

\section{Results}

\begin{figure*}
    \centering
    \includegraphics[width=18cm,height=6cm]{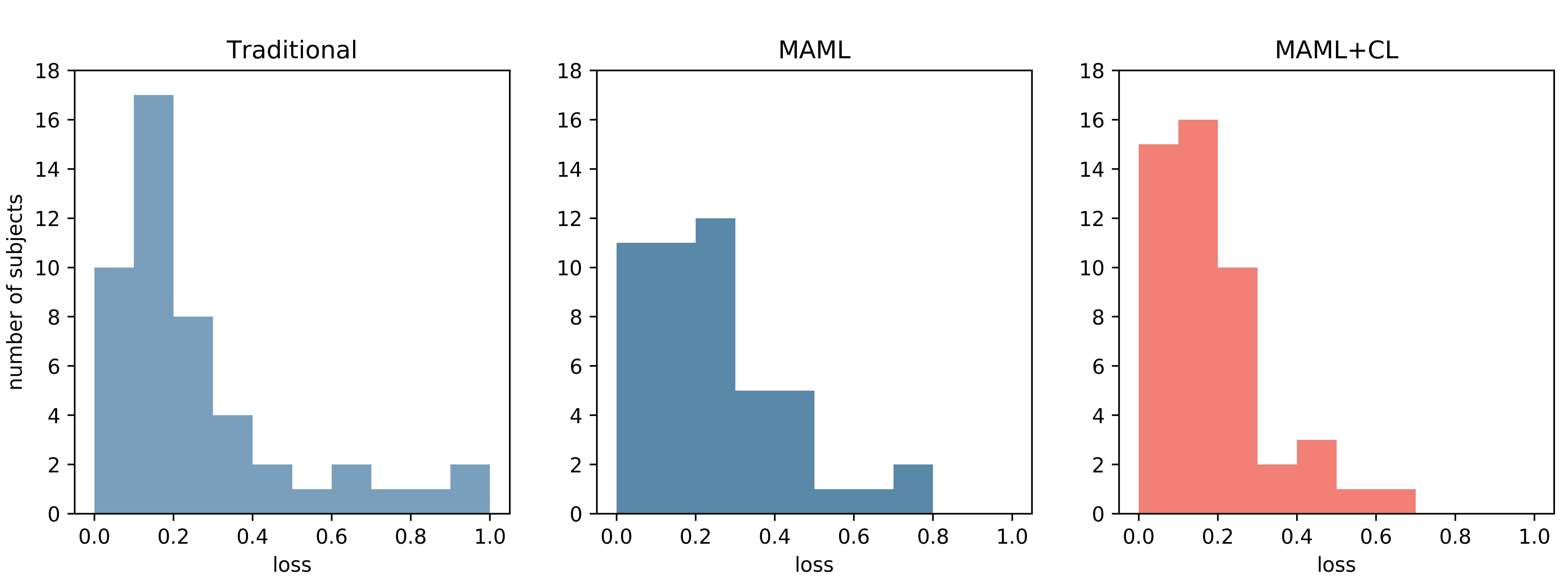}
    \caption{The number of pre-trained subjects corresponding to the losses.}
    \label{fig:result1}
\end{figure*}

\subsection{Advantages of MAML}

\autoref{table:results1} shows the results of the compared methods on four metrics. We can see that weights $\theta_0$ trained by MAML or the MAML+CL-selector outperform the traditionally trained weights. Regardless of the pre-trained weights $\theta_0$ being fine-tuned using any method, the performance of MAML is obviously better than traditional pre-training. The weights pre-trained by simple MAML and traditional pre-training yield an ROC-AUC of 0.9703 and 0.9566 , respectively, by using the same fine-tuning. Rather than comparing the precision and recall separately, we prefer to compare the F1-score, which combines the two metrics. The F1-score of the three weights under traditional fine-tuning reaches 0.9064, 0.8902, and 0.8833. The reason why MAML outperforms traditional method is that the MAML weights may have relatively balanced performance on most subjects, where the traditional pre-training weights performed unsatisfactorily when faced with some subjects. This unbalanced phenomenon occurs even on the pre-trained subjects (see \autoref{fig:result1}). 
We calculate the losses of pre-trained parameters by using MAML+CL, MAML, and the traditional approach on pre-trained tasks five times and normalize the losses by the maximum loss. Then, we divide the losses into intervals of 0.1 and count the number of pre-training subjects whose loss is in the intervals. It is obvious that traditionally pre-trained weights perform poorly on more subjects. By conducting a paired T test, we assert that MAML can help adapt to different subjects because the p-value is $1.02\times 10^{-7}$, much less than 0.01.

\subsection{Help of CL-selector}

We observe that the CL-selector can gain better weights $\theta_0$ than simple MAML, as shown in \autoref{table:results1}. The PR-AUC shows the largest difference between using and not using the CL-selector in both fine-tuning methods. The MAML+CL-selector achieves 0.9681 when fine-tuned with our method and 0.9512 when fine-tuned with simple fine-tuning; whereas simple MAML obtains relatively worse results: 0.9588 and 0.9423, respectively. As described in previous subsection, we count the subjects according to the losses of weights pre-trained by MAML and the MAML+CL-selector to evaluate the CL-selector. As shown in \autoref{fig:result1}, our pre-trained method achieves a more balanced performance on various subjects compared with simple MAML. The p-value of the one-sided paired T test is 0.033, indicating that the CL-selector can improve MAML. This result supports our hypothesis that the CL-selector could help MAML train the model in a more appropriate order.

\subsection{Better Fine-tuning}
\begin{figure}[h]
    \includegraphics[width=\linewidth]{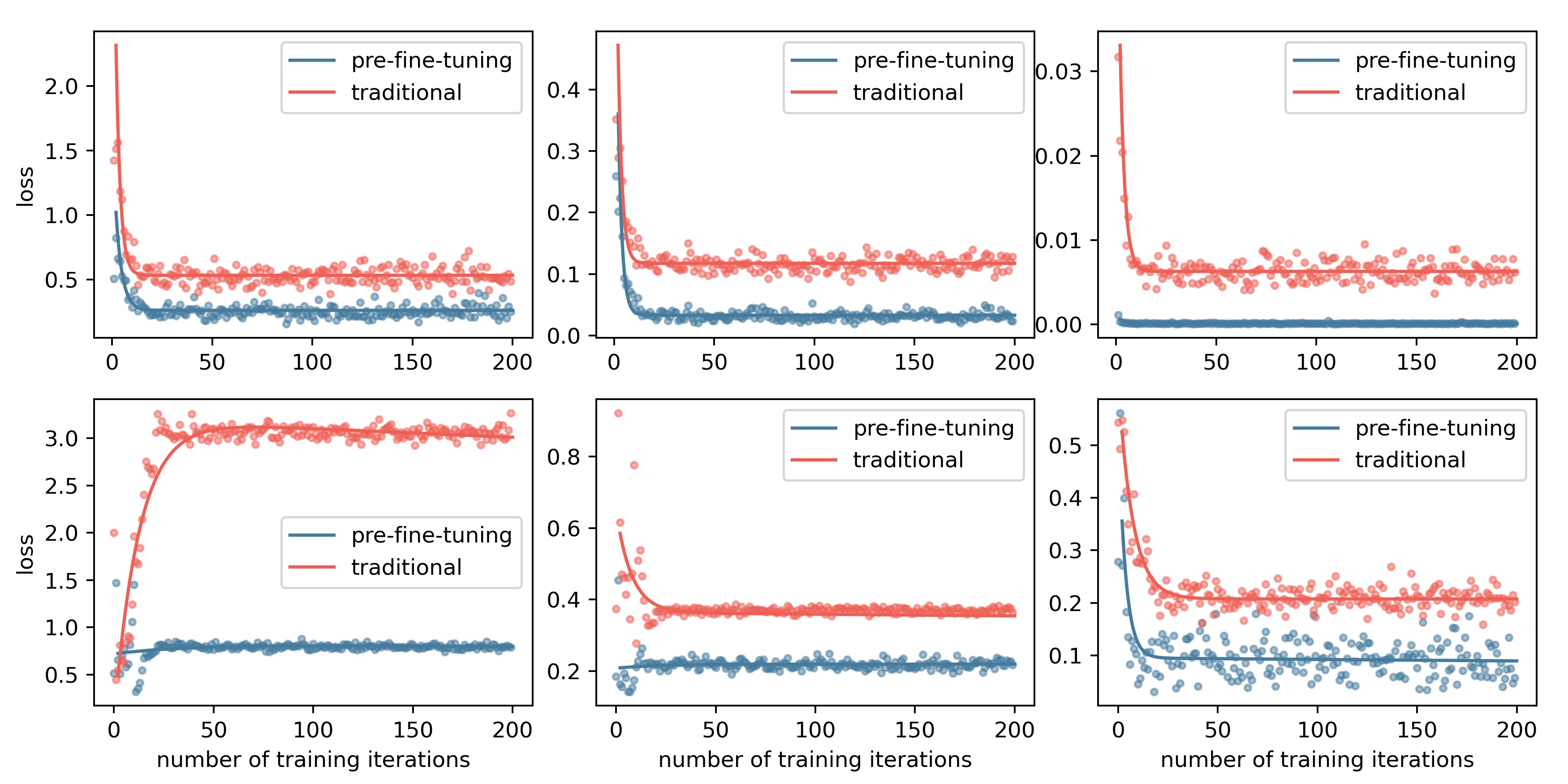}
    \caption{Losses on the training set and test set of three hardest-to-adapt subjects. The three figures above are training losses, and the three figures below are test losses. In all figures, the curves above are all for pre-fine-tuning.} 
    \label{fig:loss}
\end{figure}

Our pre-fine-tuning stage also contributes to adapting to new subjects. From \autoref{table:results1}, we can see that our fine-tuning almost improves the performance in terms of all metrics. Pre-trained and fine-tuned with our method, the model yields an ROC-AUC of 0.9843, an accuracy of 0.9602, and a PR-AUC of 0.9681, and the F1-score of our method is also the highest, reaching 0.9173. MAML combined with our pre-fine-tuning also outperforms significantly, obtaining an ROC-AUC, an accuracy, a PR-AUC, and an F1-score of 0.9823, 0.9572, 0.9588, and 0.9156, respectively. However, the MAML+CL-selector with simple fine-tuning can only reach 0.9742, 0.9549, 0.9512, and 0.9064, respectively. Consequently, this pre-fine-tuning contributes the most to our results to some extent. This is because our pre-fine-tuning can help the model fit the training set in the fine-tuning stage, particularly those subjects that are difficult to adapt to. \autoref{fig:loss} shows the training loss curves and test loss curves for the three subjects that are the most difficult to fit (with lowest ROC-AUCs). The training losses and test losses are all relatively small with our proposed pre-fine-tuning method.  

\section{Discussions and Limitations}
The results show that our \mname model can outperform the traditional method and that the CL-selector and pre-fine-tuning can improve the performance. Based on \autoref{table:results1}, we can see the improvement in our proposed methods. It is confirmed that our pre-training method can prompt the generalization ability between individuals, and our pre-fine-tuning method can help the model better fit the unseen subjects and different stages. From the results, we can see that MAML and pre-fine-tuning help to improve the metrics significantly, whereas the CL-selector helps relatively little. However, our proposed methods can bring overall improvements in terms of the metrics we used.

We observe the following: 1) meta-learning attempts to find the most potential weights (i.e., minimizing the loss after adaptation to new tasks) and not the currently optimal weights. Due to the difference between individuals, we can consider the disease detection in different subjects as various tasks, which the meta-learning methods do well in. 2) Unlike the simple MAML, the CL strategy can make the model learn ``from easy to hard''; this has been proven useful in the sample aspect \cite{09CurriculumLearning}. 3) The method of pre-fine-tuning absorbs the thought of MAML, aiming to solve the problem that recordings from identical individuals, but in different stages, could be significantly different. We can consider this procedure as the application of MAML on the sample aspect to improve the generalization ability on different samples. Although the results support our \mname model, some limitations remaain to be addressed.

Most deep learning methods focus on extracting features or architecture of the neural networks to detect VA \cite{2008Discrimination, FNN, ganresnet, jia2020personalized}. The networks are adjusted to better fit the ECG data, and better and better results are obtained. For example, Jia et al. tried to improve the ability to adapt the model to different individuals, which is to some extent similar to our goal \cite{jia2020personalized}. However, they mainly tried to adjust the network and find a better architecture to achieve better performance, whereas we attempted to find a general model-agnostic method that can improve the performance of all networks. Thus, our method could be combined with related works focusing on architectures to achieve a better and more balanced performance on different individuals. Note that MetaSleepLearner \cite{metasleeplearner} is a similar previous work. Although we have the same goal of fast adaption to different individuals for sleep staging (MetaSleepLearner) and VA detection (ours), we differ from it by further improving the pre-training stage by incorporating CL and improving the fine-tuning stage by advanced pre-fine-tuning. 

Our proposed method has a better generalization ability and is more suitable for specific requirements with some additional adjustments. First, the MAML part could be replaced by newer editions (such as Reptile \cite{Reptile}) to reduce the amount of computation without losing much accuracy. This is because the second-order derivation is replaced by an approximate one-order derivation. Based on this, some new neural networks (such as \cite{jia2020personalized}) can be pre-trained using MAML without much training time. Second, our CL-selector is a not a refined method to implement the CL strategy. We can adopt more suitable CL methods to better mine the CL ability. Two main classes of methods can be introduced: increasing the capacity of the model \cite{paper42,paper43,paper44} and improving the complexity of tasks (similar to the method we used) \cite{paper45,paper47,paper49}. Finally, our method could be followed by a post-process procedure. For example, some hypothesis testing methods can be applied to long time-series data, using predictions from several epochs. In other words, if we have consecutive VA segments, we can divide it into several epochs and use the result from all epochs to infer cooperatively.

Our method has some limitations. First, although MAML has been proven theoretically by Finn \cite{MLearningThe}, the limited theoretical proof restricts CL and our pre-fine-tuning. Some researchers have proposed learning strategies that are contrary to CL (such as anti-curriculum \cite{Anti-Curriculum}) and achieved good results. In addition, our pre-fine-tuning, similar to MAML, remains a pilot approach without any theoretical grounding. Second, the high computational cost of MAML (because of its second-order derivation) prevents it from training large-scale neural networks. Third, it is unknown whether our method could be used clinically because data in real-world scenarios may be dirty, which means that there may be incorrectly labeled data. Sometimes, when trained with such data, the model may confuse and easily over-fit a training set \cite{2016Understanding}. Because of these limitations, there is still much work to be performed in the future.

\section{Conclusion}
In this paper, we propose \mname, a VA detector that aims to transfer initial parameters that are easily generalized to new individuals to achieve fast and accurate adaptation to new individuals. Furthermore, we modified the simple MAML with a CL-selector to afford a more meaningful order of selecting the meta-tasks. Finally, we introduced the concept of MAML that trains the model for a potential parameter into fine-tuning, proposing a method called pre-fine-tuning. Three publicly available datasets were used to perform the experiments, namely, the MITDB, CUDB, and VFDB. We employed a previous state-of-the-art deep neural network architecture, pre-trained using three methods and transferred the weights to unseen individuals, followed by two fine-tuning methods. When fine-tuned, the model was fed with only 10 recordings per class to verify the ability of fast adaptation. The procedure was repeated five times in both the pre-training and fine-tuning stages to make the results more convincing and to avoid the contingency. Finally, our \mname model statistically outperformed the other combinations of methods in pre-training and fine-tuning. Four evaluation indexes were used, and we discarded precision and recall because the F1-score is more scientifically correct and treated comprehensively. Our method outperformed the others on all four measurements. Our proposed method indicates the possibility of better generalization to different patients in real-world VA detection and quickly adapting to new individuals, although there are still some restrictions remaining.

\section*{Acknowledgment}

This work was supported by the National Natural Science Foundation of China (No.62102008), and the Fundamental Research Funds for the Central Universities. 
   
\bibliographystyle{IEEEtran}
\bibliography{main}

\end{document}